\documentclass[a4paper]{svproc}

\usepackage{amsmath,amsfonts,amssymb}

\usepackage{graphicx}
\graphicspath{{./figure/}}

\usepackage{url}

\usepackage{booktabs}

\usepackage{algorithm}
\usepackage[noend]{algpseudocode}

\usepackage{cite}

\usepackage{xspace}

\usepackage{ifthen}

\usepackage{comment}

\usepackage{placeins}

\usepackage{hyperref}

\newcommand\abbrMAPF{MAPF\xspace}
\newcommand\abbrPAMO{PAMO\xspace}
\newcommand\abbrMAPFMO{M-PAMO\xspace}
\newcommand\abbrCBS{CBS\xspace}
\newcommand\abbrOursh{CBS-MOH\xspace}
\newcommand\abbrOursl{CBS-MOL\xspace}

\newboolean{shortver}
\setboolean{shortver}{true}  

\begin{document}
\mainmatter

\title{Conflict-Based Search for Multi-Agent Path Finding Among Movable Obstacles}

\titlerunning{CBS for Multi-Agent Path Finding Among Movable Obstacles}

\author{
Shaoli Hu \inst{1}\fnmsep \inst{2} \and 
Shizhe Zhao \inst{1} \and 
Zhongqiang Ren \inst{1}\thanks{Corresponding author.}\footnotetext{
This work was supported by the Natural Science Foundation of Shanghai under Grant 24ZR1435900.}
}

\authorrunning{Shaoli Hu et al.}

\tocauthor{Shaoli Hu, Shizhe Zhao, Zhongqiang Ren}

\institute{
Shanghai Jiao Tong University, Shanghai, China \and
The Hong Kong University of Science and Technology, Hong Kong, China\\
\email{zhongqiang.ren@sjtu.edu.cn}
}

\maketitle

\begin{abstract}
	This paper investigates Multi-Agent Path Finding Among Movable Obstacles (\abbrMAPFMO), which seeks collision-free paths for multiple agents from their start to goal locations among static and movable obstacles.
\abbrMAPFMO arises in logistics and warehouses where mobile robots are among unexpected movable objects.
Although Multi-Agent Path Finding (\abbrMAPF) and single-agent Path planning Among Movable Obstacles (\abbrPAMO) were both studied, \abbrMAPFMO remains under-explored.
Movable obstacles lead to new fundamental challenges as the state space, which includes both agents and movable obstacles, grows exponentially with respect to the number of agents and movable obstacles.
In particular, movable obstacles often closely couple agents together spatially and temporally.
This paper makes a first attempt to adapt and fuse the popular Conflict-Based Search (CBS) for \abbrMAPF, and a recent single-agent PAMO planner called PAMO*, together to address \abbrMAPFMO.
We compare their performance with up to 20 agents and hundreds of movable obstacles, and show the pros and cons of these approaches.
Finally, we deploy our planner on real robots among movable obstacles in a lab setting, where the robots can push the movable obstacles in correct orders to reach their respective goals, which is otherwise hard to achieve with existing methods.

\end{abstract}


\section{Introduction}\label{dms:sec:intro}

Multi-Agent Path Finding (\abbrMAPF) seeks collision-free paths for multiple agents among static obstacles from their start to goal locations while minimizing their arrival times.
This paper considers a variant called Multi-agent Path finding Among Movable Obstacles (\abbrMAPFMO), where some of the obstacles can be pushed by the agents.
When there are no movable obstacles, \abbrMAPFMO becomes \abbrMAPF.
\abbrMAPF is NP-hard~\cite{yu2013structure_nphard} and so is \abbrMAPFMO.
When there is only one agent, \abbrMAPFMO becomes the Path planning Among Movable Obstacles (\abbrPAMO)~\cite{2025_ICRA_PAMO,demaine2000pushpush,stilman2008planning}.
\abbrMAPFMO arises in logistics and warehouses where the passages are blocked by unexpected movable items or shelves~\cite{10.24963/ijcai.2023/611,vainshtein2023terraforming}.

\begin{figure}[tb]
\centering
\includegraphics[width=0.85\linewidth]{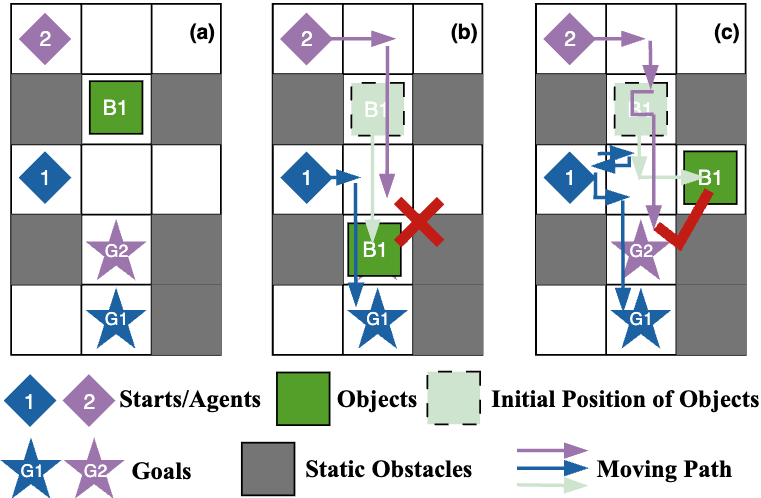}
\caption{An example of M-PAMO.
(a) shows the problem instance.
(b) shows that if agent 1 follows its shortest path and reaches its goal at time $t=3$, there is no way for agent 2 to reach its goal due to the movable obstacle $B1$, since it cannot push $B1$ onto agent 1 at time $t=4$.
(c) shows the solution returned by our \abbrOursh and \abbrOursl, where agent 1 has to wait for 2 steps at first and then pushes $B1$ to the right so that both agents can then reach their goals.}
\label{PAMO:fig:example}
\vspace{-5mm}
\end{figure}

Although \abbrMAPF~\cite{yu2013structure_nphard,sharon2015conflict} and \abbrPAMO~\cite{2025_ICRA_PAMO,demaine2000pushpush,stilman2008planning} were both studied, we are not aware of any study on \abbrMAPFMO.
The introduction of movable obstacles (referred to as objects or boxes hereafter) into \abbrMAPF leads to new fundamental challenges, and makes some of the popular techniques for \abbrMAPF less effective.
The movable obstacles lead to a much larger state space that grows exponentially with respect to the number of both the agents and the objects.
In particular, these objects often closely couple the agents together spatially and temporally (Fig.~\ref{PAMO:fig:example}): for example, an object that was pushed to location $v$ by agent $i$ at time $t$ may block the only path of agent $j$ at time $t'$ that is much later than $t$, forcing the planner to consider different ways of interaction among agents $i,j$ and the objects, to find this only solution.

As a first attempt, this paper focuses on Conflict-Based Search (\abbrCBS)~\cite{sharon2015conflict}, a popular method for \abbrMAPF, and investigates how to extend it to handle \abbrMAPFMO.
\abbrCBS represents the workspace as a graph and employs a two-level search to handle \abbrMAPF, where the low-level addresses single-agent planning and the high-level resolves agent-agent conflicts.
\abbrCBS begins by planning an individually optimal path for each agent while ignoring others, and then detects collisions among the agents along their paths.
For a conflict between agents $i,j$, the high-level of \abbrCBS resolves it by adding constraints to either $i$ or $j$ and invokes the low-level to replan agent $i$'s or $j$'s path to satisfy the added constraints.
\abbrCBS repeats this process, alternating between high-level and low-level, until a set of conflict-free paths is found.

We propose two variants of \abbrCBS for \abbrMAPFMO, where the first one \abbrOursh (Movable Obstacles on High-level) addresses the objects only at its high-level, while the second one \abbrOursl also modifies the low-level to handle the objects.
Specifically, \abbrOursh ignores objects at its low-level and can leverage any existing single-agent planners used in \abbrCBS.
\abbrOursh handles objects by introducing new types of conflicts and constraints into the search, and adds those constraints to the agents when conflicts of agent-agent, agent-object, object-object and object-environment are detected.
In contrast, \abbrOursl further considers the objects at its low-level, where each agent's path is always planned among all objects.
For this purpose, we extend our prior work for (single-agent) PAMO called PAMO*~\cite{2025_ICRA_PAMO} to plan in space-time, and name the resulting planner ST-PAMO*, which is used as the low-level of \abbrOursl.

Although \abbrCBS is guaranteed to find an optimal solution for \abbrMAPF, we show that our \abbrOursh and \abbrOursl are not guaranteed to find optimal solutions for \abbrMAPFMO by providing intuitive counter-examples.
The development of \abbrOursh and \abbrOursl and the examples show the algorithmic challenge in \abbrMAPFMO, which provides clues for future research.
We compare two proposed algorithms in various maps with up to 20 agents and hundreds of objects, which demonstrate the reasonable scalability of our methods, in terms of both the number of agents and the number of movable obstacles.

Finally, we also deploy our planner on real robots among movable obstacles in a lab setting.
In this real-robot demo, the robots have to push the movable obstacles in certain order with careful collaboration to navigate towards their respective goal, which is otherwise hard to achieve either with the existing \abbrMAPF methods that ignore movable obstacles, or with the existing single-agent \abbrPAMO planner that ignores the robot-robot interaction and collaboration.

\subsection{Related Work}\label{dms:sec:related}

\abbrMAPF algorithms fall on a spectrum from coupled~\cite{standley2010finding} to decoupled~\cite{silver2005cooperative}, trading off completeness and optimality for scalability.
In the middle of this spectrum lies the popular dynamically coupled methods such as the popular CBS \cite{sharon2015conflict}, which has been extended and improved in many ways~\cite{barer2014suboptimal,ren22mocbs_tase,ren23cbssTRO}.
Some recent work considers modifiable environments where the shelves in the warehouse can be rearranged when needed before planning paths for the agents~\cite{10.24963/ijcai.2023/611} or when some agents have no tasks~\cite{vainshtein2023terraforming}, which is different from \abbrMAPFMO.

PAMO was formulated and studied in different ways, such as in a grid~\cite{demaine2000pushpush}, with polygonal obstacles~\cite{wilfong1988motion}, known environments~\cite {stilman2008planning}, and unknown environments~\cite{he2024interactive}, using search~\cite{stilman2008planning}, sampling~\cite{van2010path}, and learning~\cite{xia2020interactive} methods.
However, none of them provides completeness or solution quality guarantees.
The difficulty in PAMO is that the state space includes not only the robot's position but also the objects' positions, and the dimensionality of the state space grows exponentially as the number of objects increases.
Our prior work~\cite{2025_ICRA_PAMO} presents an efficient approach called PAMO* with completeness and optimality guarantees, which runs A*-like search while leveraging the fact that most of the objects are intact during planning, which thus helps limit the actual state space explored during planning.

\section{Problem Statement}\label{dms:sec:problem}

Let the index set $I_A=\{1,2,\dots,A\}$ denote a set of $A$ agents.
All agents move in a workspace represented as a finite grid-like graph $G=(V,E)$, where $G$ is a 2D occupancy grid that represents the workspace, where each cell $v\in V$ has coordinates $v=(x,y) \in \mathbb{Z}^+$ indicating the column and row indices of the cell in the grid.
The time dimension is discretized into time steps.
At any cell $v=(x,y)$, the possible actions of an agent are to wait in place or to move in one of the four cardinal directions and arrive at one of the neighboring cells $\{(x+1,y),(x-1,y),(x,y+1),(x,y-1)\}$.
Each action takes a time step.
Let $N(v)$ denote the set of neighboring cells of $v\in V$.
Each edge $e=(v_1,v_2) \in E$ indicates the movement from cell $v_1$ to another cell $v_2$.

At any time, all cells in $V$ are partitioned into three subsets: free space $V_{free}$, static obstacles $V_{so}$, and movable obstacles (i.e., objects, or boxes) $V_{mo}$, i.e., $V = V_{free}\cup V_{so} \cup V_{mo}$.
Since the boxes can be pushed as explained later, the sets $V_{mo}$ and $V_{free}$ change over time.
The boxes are numbered by the index set $I_B=\{1,2,\cdots,|V_{mo}|\}$.
Each cell $v\in V_{free}$ is obstacle-free and can be occupied by an agent.
Each cell $v\in V_{so}$ represents a static obstacle and cannot be occupied by any agent at any time.
Each cell $v\in V_{mo}$ indicates a box.
When an agent is adjacent to a box $v \in V_{mo}$ and moves to cell $v$, the box at $v$ is pushed in the same direction as the agent's movement simultaneously.
Specifically, a box $v \in V_{mo}$ can be pushed by an agent to an adjacent cell if all of the following conditions hold: (i) the current cell $v_c$ occupied by an agent $i\in I_A$ is next to $v$; (ii) the agent $i\in I_A$ takes an action to reach $v$; and (iii) the box can be pushed into a neighboring cell $v' = v + (v_c-v)$ that is in free space (i.e., $v' \in V_{free}$) and is not occupied by another agent $j\in I_A, j\neq i$.
Note that, Condition (iii) ensures a box cannot be pushed to $v$ if $v$ is occupied by another box or agent.
When a box is pushed, the agent's action is also referred to as a push.

We use a superscript $i \in I_A$ over a variable to show the agent to which the variable belongs (e.g. $v^i\in V$ is a vertex related to agent $i$).
Let $v_s^i, v^i_g \in V$ respectively denote the start and goal location of agent $i$.
All agents share a global clock and start the paths at time $t=0$.
Each action of an agent, either wait, move or push, requires one unit of time. 
Let $\pi^i(v^i_{1}, v^i_{\ell})$ be a path from $v^i_{1}$ to $v^i_{\ell}$ via a sequence of vertices $(v^i_{1},v^i_{{2}},\dots,v^i_{\ell})$ in $G$.
Each pair of adjacent vertices in $\pi^i$ corresponds to an action of the agent.
Let $g(\pi^i(v^i_{1}, v^i_\ell))$ denote the cost value of the path, which is the arrival time at $v^i_\ell$ along the path.
Here, $g(\pi^i(v^i_{1}, v^i_\ell))$ is equal to the number of actions along the path, i.e., $g(\pi^i(v^i_{1}, v^i_{\ell})) = \Sigma_{j=1,2,\dots,{\ell-1}} cost(v^i_{{j}}, v^i_{{j+1}})$, with $cost(v^i_{{j}}, v^i_{{j+1}})=1$.
We denote $\pi^i(v^i_{1}, v^i_{\ell})$ simply as $\pi^i$, and $g(\pi^i)$ as $g^i$ when there is no confusion.

Any two agents (or two boxes) are in agent-agent conflict (or box-box conflict) if one of the following two cases happens.
The first case is a vertex conflict where any two agents (or two boxes) occupy the same location at the same time.
The second case is an edge conflict where any two agents (or two boxes) move through the same edge from opposite directions between times $t$ and $t+1$ for some $t$.
Similarly, there is an agent-box conflict if an agent and a box occupy the same vertex or move through the same edge at the same time.
All of agent-agent (AA), box-box (BB), and agent-box (AB) conflicts (Fig.~\ref{PAMO:fig:AB}) are simply referred to as conflicts when there is no ambiguity.

\begin{problem}[\abbrMAPFMO]
Multi-Agent Path Finding Among Movable Obstacles (\abbrMAPFMO) seeks a set of conflict-free paths $\{\pi^i\}$ for all agents $i \in I_A$, while the sum of individual path costs $\sum_{i\in I_A} g(\pi^i)$ reaches the minimum.
\end{problem}

When $V_{mo}=\emptyset$, \abbrMAPFMO becomes \abbrMAPF. When there is only one agent $I_A=\{1\}$, \abbrMAPFMO becomes PAMO.

\section{Method}\label{dms:sec:method_definition}
\subsection{Review of Conflict-Based Search}

CBS \cite{sharon2015conflict} is a two-level algorithm.
The high-level maintains a Constraint Tree (CT).
Each node $P=(\pi,g,\Omega)$ contains a joint path $\pi = (\pi^1, \pi^2, \ldots, \pi^A)$, a cost value $g=\sum_{i\in I_A} g(\pi^i) $, and a constraint set $\Omega$.
The form of each constraint is $(i, v, t)$ or $(i, e, t)$, which means agent $i$ is prohibited from occupying vertex $v$ or traversing edge $e$ at time $t$.
CBS constructs a CT with the root node $P_{root}=(\pi_o,g(\pi_o),\emptyset)$, where the joint path $\pi_o$ is obtained by running the low-level (single-agent) planner (\texttt{LowLevelPlan}), such as A*, for every agent respectively with an empty set of constraints while ignoring any other agents.
$P_{root}$ is added to OPEN, a queue that prioritizes nodes based on their $g$-values from the minimum to the maximum.

In each iteration, a node $P=(\pi,g,\Omega)$ with the minimum $g$-value is selected from OPEN for expansion, where every pair of paths in $\pi$ is checked for a vertex conflict $(i,j,v,t)$ (and an edge conflict $(i,j,e,t)$).
If no conflict is detected, $\pi$ is conflict-free and is returned as a \emph{solution} (i.e., a conflict-free joint path) and this solution is guaranteed to have the optimal cost.
Otherwise, two constraints $\omega^i =(i,v,t)$ and $\omega^j = (j,v,t)$ are generated respectively, and two new constraint sets $\Omega\bigcup\{i,v,t\}$ and $\Omega\bigcup\{j,v,t\}$ are generated.
Edge conflict is handled in a similar manner.
For agent $i$ in each of those two constraints $(i,v,t)$ and the corresponding newly generated constraint set $\Omega'=\Omega\bigcup\{i,v,t\}$, the low-level planner is invoked to plan a minimum cost path ${\pi'}^i$ of agent $i$ subject to all constraints related to agent $i$ in $\Omega'$.
The low-level planner typically runs A*-like search in a space-time graph with constraints marked as inaccessible vertices and edges in this space-time graph.
After the low-level search, a new joint path $\pi'$ is formed by first copying $\pi$ and then replacing agent $i$'s path $\pi^i$ with $\pi'^i$.
Finally, for each of the two generated constraints, a corresponding new node is generated and added to OPEN for future expansion.
CBS~\cite{sharon2015conflict} is guaranteed to find a minimum cost solution for a solvable \abbrMAPF.

\subsection{Resolving AB, BB and BPR Conflicts}\label{M_PAMO:sec:resolve_conflict}

\begin{figure}[tb]
\centering
\includegraphics[width=0.8\linewidth]{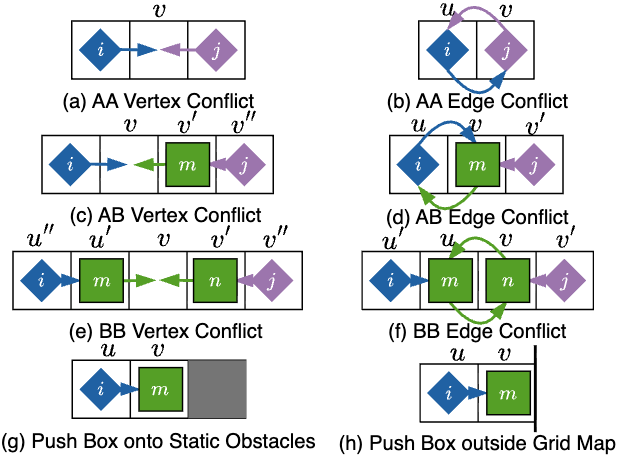}
\caption{Different types of conflicts in \abbrMAPFMO.}
\label{PAMO:fig:AB}
\end{figure}

We first present the conflict resolution approach for \abbrMAPFMO, which is used by both \abbrOursh and \abbrOursl.
Among the three types of conflicts, AA conflicts are resolved in the same way as in CBS.
We now focus on resolving AB and BB conflicts.
Boxes cannot move themselves and can only be pushed by an agent.
To prevent a box from occupying a vertex, we forbid the corresponding push action of the agent that moves the box by adding an edge constraint.
Specifically, for an AB vertex conflict $(i, m, v, t)$ as shown in Fig.~\ref{PAMO:fig:AB} (c), where agent $i \in I_A$ and box $m \in I_B$ occupy the same vertex $v$ at time $t$, the following two constraints are generated to resolve the conflict:
\begin{itemize}
    \item Constraint $(i,v,t)$ prevents agent $i$ from occupying $v$ at time $t$.
    \item Constraint $(j,(v'',v'),t)$ forbids agent $j\in I_A$ to move from $v''$ to $v'$ ($v'', v' \in V_{free}$) between times $t-1$ and $t$, which pushes the box $m$ from $v'$ to $v$ between times $t-1$ and $t$.
\end{itemize}
Note that, the second constraint $(j,(v'',v'),t)$ is imposed on the agent $j$ that pushed box $m$ instead of adding constraint on the box itself.
We discuss in Sec.~\ref{M_PAMO:sec:discuss} why not add constraints onto the boxes.

An AB edge conflict $(i, m, (u, v), t)$ can be resolved in a similar manner as AB vertex conflict. 
As shown in Fig.~\ref{PAMO:fig:AB} (d), agent $i \in I_A$ and box $m \in I_B$ swap their locations over the same edge $(u,v)$ (i.e., $u \in V_{free}$) at time $t$.
The following two constraints are generated to resolve the conflict: Constraint $(i,(u,v),t)$ prevents agent $i$ from $u$ to $v$ at time $t$, and constraint $(j,(v',v),t)$ forbids agent $j$ to move from $v'$ to $v$ between times $t-1$ and $t$, which pushes the box $m$ from $v$ to $u$ between times $t-1$ and $t$.

For a BB vertex conflict $(m, n, v, t)$ as shown in Fig.~\ref{PAMO:fig:AB} (e), where box $m \in I_B$ and box $n \in I_B$ occupy the same vertex $v$ at time $t$, the following two constraints are generated to resolve the conflict: Constraint $(i,(u'',u'),t)$ forbids agent $i\in I_A$ to move from $u''$ to $u'$ ($u'',u' \in V_{free}$) between times $t-1$ and $t$, which pushes the box $m$ from $u'$ to $v$ between times $t-1$ and $t$. Constraint $(j,(v'',v'),t)$ forbids agent $j\in I_A$ to move from $v''$ to $v'$ between times $t-1$ and $t$, which pushes the box $m$ from $v'$ to $v$ between times $t-1$ and $t$.
A BB edge conflict $(m, n,(u, v), t)$, as shown in Fig.~\ref{PAMO:fig:AB} (f), can be resolved similarly by the following two constraints $(i,(u',u),t)$ and $(j,(v',v),t)$.

Finally, our approach also needs to check the situations where a box is pushed onto a static obstacle (Fig.~\ref{PAMO:fig:AB} (g)) or pushed outside the grid map (Fig.~\ref{PAMO:fig:AB} (h)).
We refer to these two situations as \emph{box pushing rules (BPR)}.
If a push action of an agent $i$ leads to violation of the BPR, a corresponding constraint $(i,e,t)$ is generated to forbid agent $i$ to use the edge $e$ (that is, perform that push action) at time $t$.
The violation of BPR yields only one constraint on agent $i$, which differs from the two constraints generated from an AA, AB or BB conflict.
Such BPR checks are needed since the low-level of \abbrOursh (next section) does not consider any box and may return a path that pushes a box outside the map.

\subsection{\abbrOursh: Handling Objects on the High-Level}

The \abbrOursh algorithm is similar to CBS.
The major difference lies in conflict detection (Alg.~\ref{alg:MOH&MOL} Line \ref{MOH&MOL:line:Sim}).
While CBS checks for AA conflicts along a joint path, \abbrOursh simulates the joint path of the boxes $\pi^{I_B}=\{\pi^m | m\in I_B\}$ based on the agents' joint path $\pi=\{\pi^i,i\in I_A\}$, and checks for conflicts and violation of BPR.

To check for conflicts ($cft$) or violation of BPR ($bpr$), as shown in Alg.~\ref{alg:box-trajectory-conflict}, \texttt{SimAndConflictCheck} initializes an empty joint path for boxes $\pi^{I_B}$ and then sets the initial boxes' positions $\pi^{I_B}(t=0)$ as in $V_{mo}$.
Then, it gets the maximum time $t_{max}$ of the longest agent path $\pi^i\in\pi,i\in I_A$, which is used as the number of simulation steps.
For each time step $t = 1,2,\cdots,t_{max}$, the positions of the boxes at time $t$ can be obtained based on the positions of the agents at times $t - 1$ and $t$ and the positions of the boxes at time $t-1$.
Meanwhile, AA, AB, and BB conflicts as well as the violation of BPR are detected for each time step.
If any conflict $cft$ or $bpr$ is detected {by Alg.~\ref{alg:box-trajectory-conflict} \texttt{Check}}, that $cft$ or $bpr$ is immediately returned to Alg. ~\ref{alg:MOH&MOL} for constraint generation ({\texttt{GenerateConstraint}}).
Again, if a $bpr$ is detected (i.e., $cft=$NULL, $bpr\neq$NULL), \texttt{GenerateConstraints} returns only one constraint (Alg.~\ref{alg:MOH&MOL} Line~\ref{MOH&MOL:line:GenConstraint}).
If neither $cft$ nor $bpr$ are found, \texttt{SimAndConflictCheck} returns NULL and terminates, indicating that the given joint path $\pi$ of the agents is conflict-free.

\FloatBarrier
\begin{algorithm}[tb]
    \caption{High-level of \abbrOursh \& \abbrOursl}
    \label{alg:MOH&MOL}
    \begin{algorithmic}[1]
        \small
        \State $\Omega_0 \gets \emptyset$
        \State $\pi_o,g_o \gets \texttt{LowLevelPlan}(\Omega_0)$ 
        \State Add $P_{root}=(\pi_o,g(\pi_o),\emptyset)$ to OPEN
        \While{OPEN not empty}
            \State $P_k = (\pi_k,g_k,\Omega_k) \gets$ OPEN.pop()
            \State $cft,bpr \gets \texttt{SimAndConflictCheck}(\pi_k)$ \label{MOH&MOL:line:Sim}
            \If{$cft$ and $bpr$ is NULL}
                \State \Return $\pi_k$
            \EndIf
            \State $\Omega \gets \texttt{GenerateConstraints}(cft,bpr)$
            \label{MOH&MOL:line:GenConstraint}
            \For{each $\omega^a\in \Omega$ }
                \State $\Omega_k' = \Omega_k\bigcup\{\omega^{a}\} $
                \State $\pi_k',g_k' \gets \texttt{LowLevelPlan}(a,\Omega_k')$
                \State Add $P_{k'} = (\pi_k',g_k',\Omega_k')$ to OPEN\label{alg:cbsmoh:lineAddOpen}
            \EndFor
        \EndWhile
        \State \Return failure
    \end{algorithmic}
\end{algorithm}

\vspace{-26pt}

\begin{algorithm}[tb]
    \caption{\texttt{SimAndConflictCheck}}
    \label{alg:box-trajectory-conflict}
    \begin{algorithmic}[1]
        \small
        \State $\pi^{I_B} \gets \emptyset$
        \State $\pi^{I_B}(t=0) \gets V_{mo}$
        \State $t_{max} \gets \max_{i\in I_A} g(\pi^i)$
        \For{$t = 1,2,\cdots,t_{max}$}
        \State $\pi^{I_B}(t) \gets \texttt{SimStep}(\pi^{I_B}(t-1), \pi(t-1), \pi(t))$
        \State $cft,bpr \gets \texttt{Check}(\pi^{I_B}(t-1), \pi^{I_B}(t), \pi(t-1), \pi(t)))$
        \If{$cft \neq$ NULL \textbf{or} $bpr \neq$ NULL}
        \State \Return $cft,bpr$
        \EndIf
        \EndFor
        \State \Return NULL
        
    \end{algorithmic}
\end{algorithm}
\FloatBarrier

\subsection{\abbrOursl: Handling Objects on the Low-Level}

The high-level of CBS-MOL is the same as that of \abbrOursh.
The biggest difference lies in their low-levels.
We first extend PAMO*~\cite{2025_ICRA_PAMO} to plan in the space-time and then use it as the low-level planner of \abbrOursl to handle constraints added by the high-level.

PAMO* addresses the (single-agent) \abbrPAMO problem using A*-like search in a state space $S=G^{1+|V_{mo}|}=G\times G\times\cdots\times G$ that encodes the position of both the agent being planned and all the boxes.
When generating successor states, PAMO* discards successors that are invalid based on the motion of the agent and the boxes.
The search is guided by a heuristic that is based on the shortest path to the goal while ignoring all the boxes.

We extend PAMO* to ST-PAMO*, which plans in a time-augmented 
state space $S\times T=G^{1+|V_{mo}|}\times T$, where $T=\{1,2,\cdots,t_h\}$ denotes the set of $t_h$ total time steps.
When generating the successors, in addition to validating agent-box motion as in PAMO*, ST-PAMO* also checks against the added constraints from the high-level search.
If any of those constraints is violated, the successor is discarded.
ST-PAMO* terminates when a path reaching the goal $v^i_g$ is found.
If any constraint $\omega$ is added at $v^i_g$, ST-PAMO* finds a path in $S\times T$ that reaches $v^i_g$ at a time step later than the time of any of those constraints $\omega$, so that the agent can stay at $v^i_g$ after the arrival.

\begin{figure}[tb]
\centering
\includegraphics[width=0.9\linewidth]{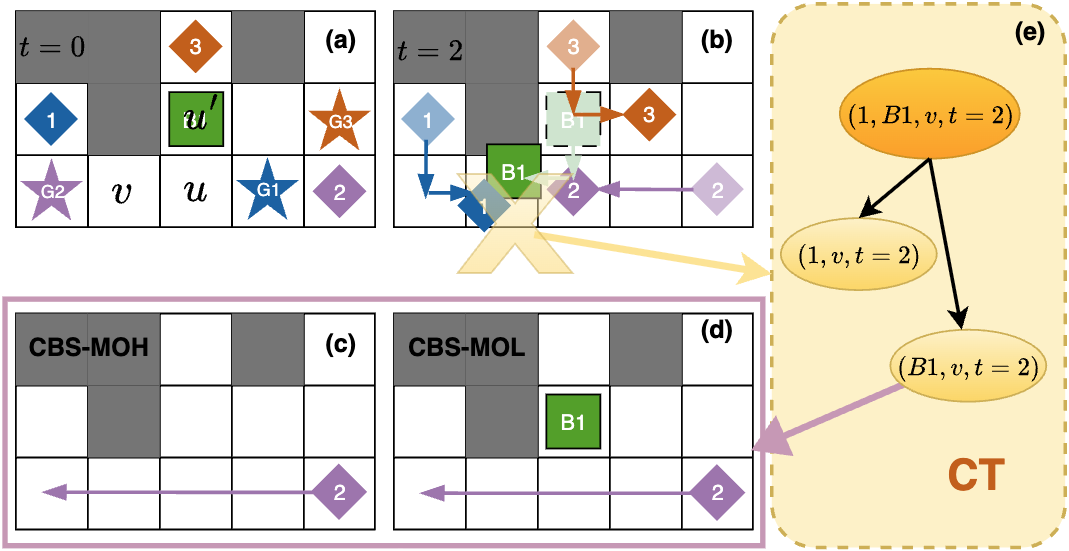}
\caption{An example where adding constraints onto the boxes fails. (a) shows the instance. (b) shows a conflict at t=2. (c) and (d) show the low-level planning: \abbrOursh ignores the box, while \abbrOursl still regards the box at its original position. (e) shows the constraint tree where the added constraint fails to resolve the conflicts and the search never terminates. More details can be found in Sec. \ref{M_PAMO:sec:discuss}.}
\label{PAMO:fig:BoxNotConsider}
\end{figure}

In \abbrOursl, each invocation of ST-PAMO* always considers the initial positions of the boxes and is not aware of the boxes that are pushed by other agents.
In other words, ST-PAMO* can only handle boxes implicitly by satisfying the constraints imposed on that agent.
For this reason, the BPR checks are still useful in \abbrOursl, since it is possible that agent $i$ pushes a box $m$ at first which blocks the path of agent $j$, and agent $j$ then pushes the box outside the map without being aware of it.
Fig.~\ref{PAMO:fig:BoxNotConsider} shows a similar situation.

\subsection{Discussion and Analysis}\label{M_PAMO:sec:discuss}

\begin{figure}[!htb]
\centering
\includegraphics[width=0.9\linewidth]{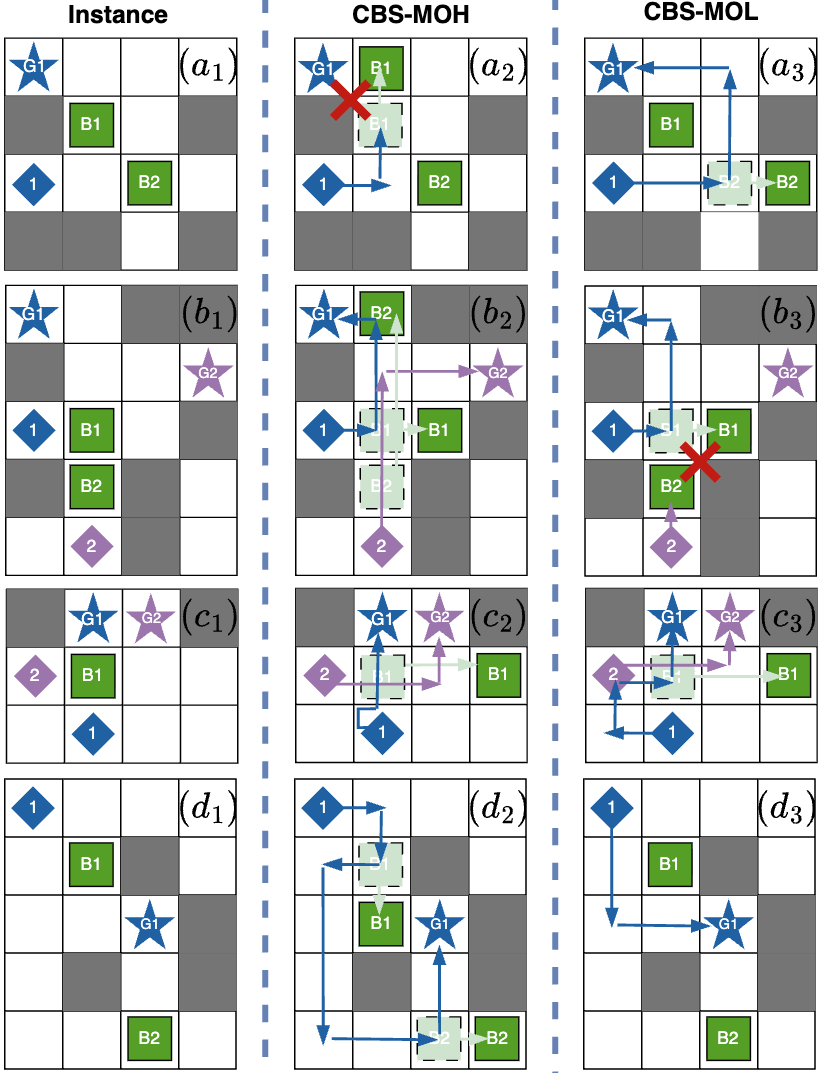}
\caption{\abbrMAPFMO examples. The first column shows 4 different instances. The 2nd and 3rd column show the planned results of \abbrOursh and \abbrOursl respectively. Row (a) shows a case where \abbrOursh fails while \abbrOursl succeeds. Row (b) shows a case where \abbrOursh succeeds while \abbrOursl fails. Row (c) shows a case where both planners succeed but find different paths, and \abbrOursl fails to find an optimal solution. Row (d) shows a case where \abbrOursh fails to find an optimal solution.}
\vspace{-5mm}
\label{PAMO:fig:DiscussionHL}
\end{figure}

\subsubsection{Constraints on the Boxes}

Both our \abbrOursh and \abbrOursl add constraints on the agents to resolve AA, AB and BB conflicts, as opposed to adding constraints on the boxes.
We provide an example in Fig.~\ref{PAMO:fig:BoxNotConsider} to illustrate the consequence of adding constraints onto the boxes.
Consider the instance in Fig.~\ref{PAMO:fig:BoxNotConsider} (a), at $t=2$, there is a conflict $(1,B1,v,t=2)$ between agent $1$ and box $B1$ at vertex $v$.
Different from the proposed constraint generation as aforementioned in Sec.~\ref{M_PAMO:sec:resolve_conflict}, an alternative way is to generate constraints $\omega^1 = (1,v,t=2)$ on agent $1$ for one branch, and $\omega^2 = (B1,v,t=2)$ on the box $B1$ for the other branch.
The constraint $\omega^2$ means that, for any future nodes following this branch, no matter which agent is replanned, the low-level planner must ensure the box $B1$ is not pushed to $v$ at time $t=2$.
However, for this instance, when replanning agent $2$ after adding $\omega^2$, the low-level planner of agent $2$ is not aware that the box $B1$ is pushed to vertex $u$ at time $t=1$. Instead, the low-level planner of agent $2$ still regards $B1$ at $u'$ and the optimal path returned would still be the same.
As a result, the high-level search would detect the same conflict again and the algorithm never terminates.
Additional information, such as boxes that were pushed by other agents at different times, needs to be passed to the low-level in an intelligent way, which is left as future work.


\subsubsection{Loss of Completeness and Optimality}

The proposed \abbrOursh and \abbrOursl are neither complete nor solution optimal as examplified in Fig.~\ref{PAMO:fig:DiscussionHL}.
Fig.~\ref{PAMO:fig:DiscussionHL} ($a_2$) shows \abbrOursh fails to find a solution.
Its low-level planner find a path that first pushes $B1$ to $v_{B1'}$ and then further pushes $B1'$ outside the map, which violates BPR.
After adding the constraints, the low-level then finds a path that pushes $B1'$ from right to left, moving the box to its goal.
Fig.~\ref{PAMO:fig:DiscussionHL} ($b_3$) shows a failure case for \abbrOursl. When planning for agent $2$, since an agent cannot push two boxes simultaneously, ST-PAMO* fails to find a feasible path for agent $2$.
\abbrOursl cannot find the feasible solution that lets agent $1$ first push $B1$ to $v_{B1'}$.
Fig.~\ref{PAMO:fig:DiscussionHL} ($c_2$) shows \abbrOursh finds an optimal solution while \abbrOursl ($c_3$) is suboptimal.
\abbrOursl's low-level planner for agent $1$ does not know that $B1$ will be pushed by agent $2$ and thus plans a path with a costly, unnecessary push action. 
In contrast, the high-level of \abbrOursh resolves the AA vertex conflict, so it finds a solution where agent $1$ is constrained to wait for agent $2$ to move $B1$, resulting in a lower total cost.
Fig. ~\ref{PAMO:fig:DiscussionHL} ($d_1$) shows a case where \abbrOursh is suboptimal, while \abbrOursl is optimal. 
\abbrOursh's low-level finds a shortest path without considering boxes.
Fig.~\ref{PAMO:fig:DiscussionHL} ($d_2$) shows, in \abbrOursh, agent $1$ pushes box $B1$ to $v_{B1'}$, resulting in a suboptimal path.

\section{Experimental Results}\label{dms:sec:result}
\vspace{-3mm}
\begin{figure}[!htb]
\centering
\includegraphics[width=1\linewidth]{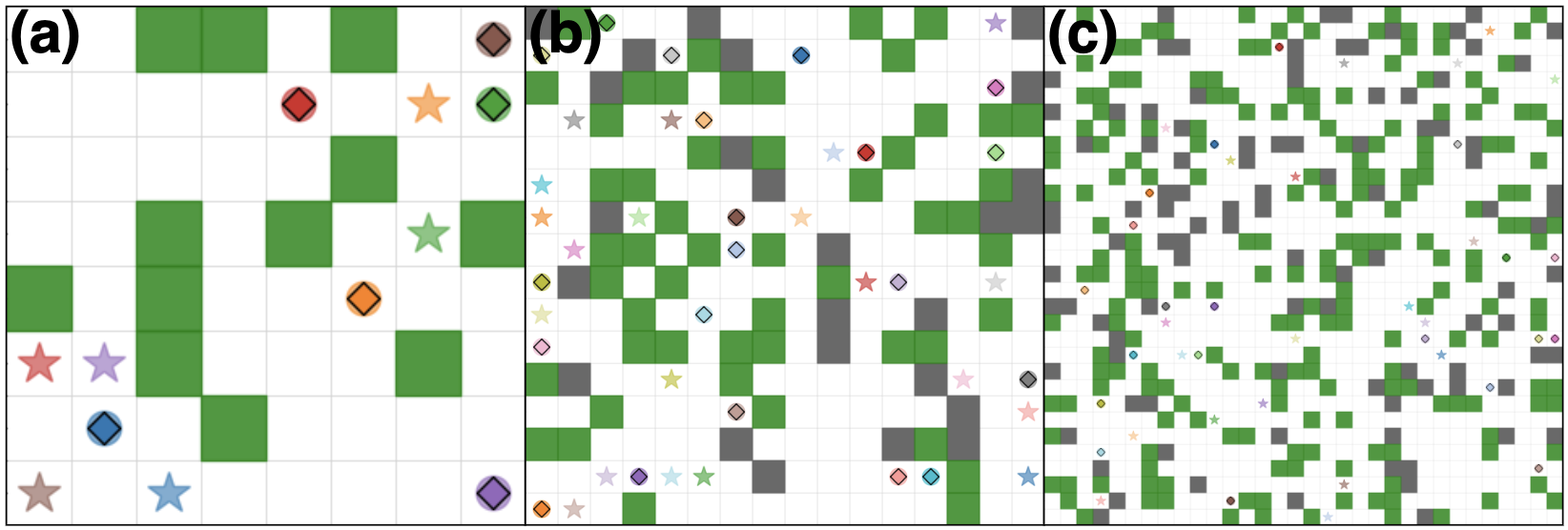}
\caption{Test instance examples. 
(a) An $8\times8$ empty grid with 6 agents, and 20\% box density; 
(b) a $16\times16$ random grid with 20 agents, 10\% static obstacles, and 20\% box density; 
and (c) a $32\times32$ random grid with the same obstacle and box densities as (b) but also 20 agents.}
\label{PAMO:fig:Grid}
\end{figure}

We test \abbrOursh and \abbrOursl in grid maps: empty $8\times 8$, $16\times 16$ and random $16\times16$, $32\times32$, as shown in Fig.~\ref{PAMO:fig:Grid}.
The empty maps have no static obstacles while the random maps have 10\% static obstacles.
We place the boxes $V_{mo}$ randomly in the grid without overlapping with static obstacles or the agents' start and goal positions.
For each map, we vary the percentage of boxes among 10\% and 20\%.
We create 25 instances for each map, and the instances are not guaranteed to be feasible due to the randomly located objects.
We set a runtime limit of 60 seconds for each instance.
All tests are conducted on a MacBook laptop with an M2 CPU and 16GB RAM.

\subsection{Success Rate}

\begin{figure}[!htb]
\centering
\includegraphics[width=1\linewidth]{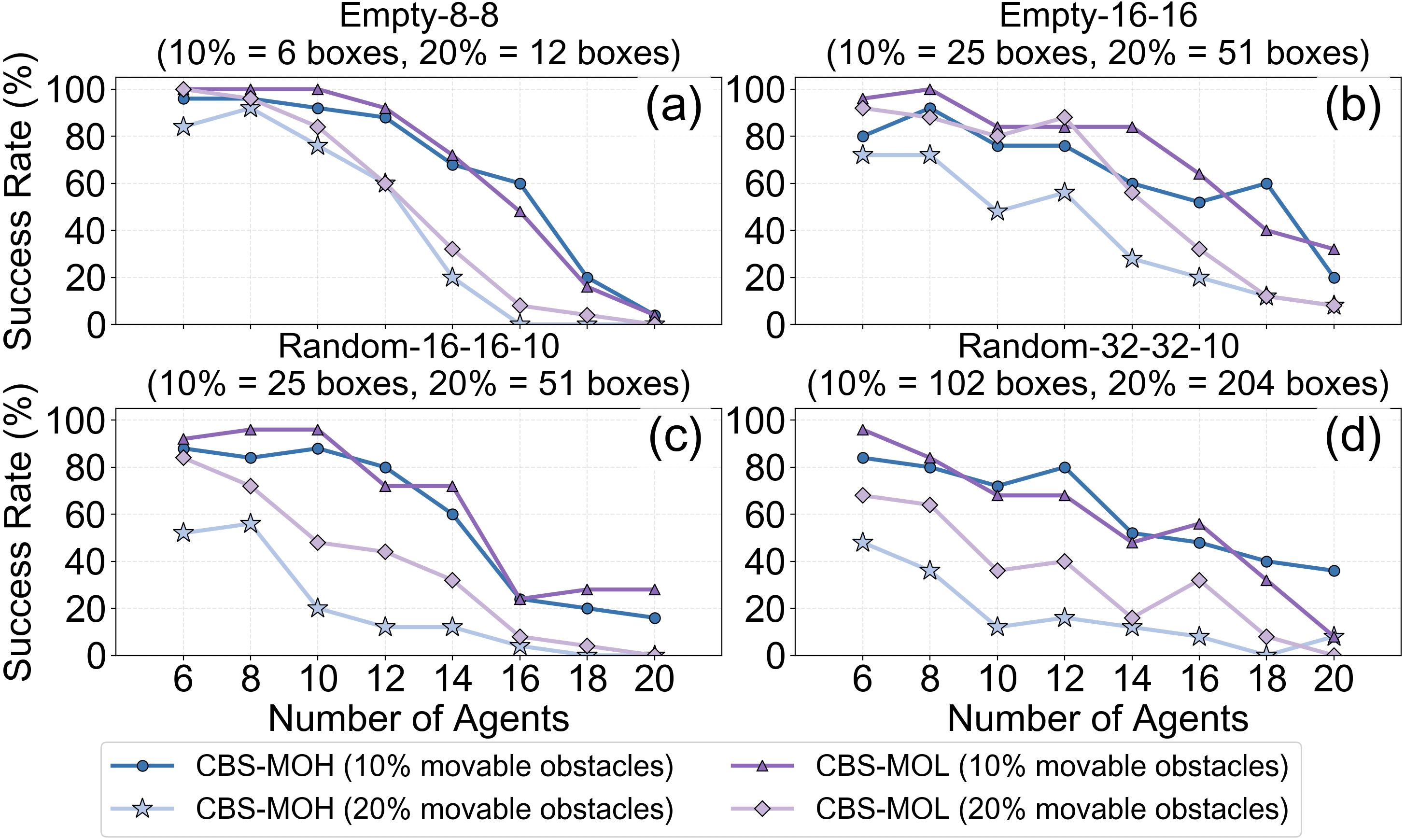}
\caption{Success rates of \abbrOursh and \abbrOursl under a varying number of boxes and a varying number of agents across different grid maps.}
\label{PAMO:fig:success}
\end{figure}

Fig.~\ref{PAMO:fig:success} shows the success rates of \abbrOursh and \abbrOursl for different numbers of agents up to $20$ in four maps, with box density varied between 10\% and 20\%.
For a fixed number of agents, both methods have lower success rates as box density increases.
The success rates of \abbrOursh drop faster than \abbrOursl, and a possible reason is the low-level ST-PAMO* in \abbrOursl considers the boxes and can handle more boxes.

For a fixed map, increasing the number of agents also reduces success rates for both methods. At 10\% box density, their performance is comparable. However, at 20\% box density, \abbrOursl consistently achieves higher success rates than \abbrOursh. This is because \abbrOursh’s low-level planner ignores boxes, leading to more deadlocks in environments  with high box density, whereas \abbrOursl’s low-level search (ST-PAMO*) explicitly accounts for boxes during planning and thereby avoids many such failures.

\FloatBarrier

\subsection{Expanded Nodes}

\begin{figure}[!htb]
\centering
\includegraphics[width=1\linewidth]{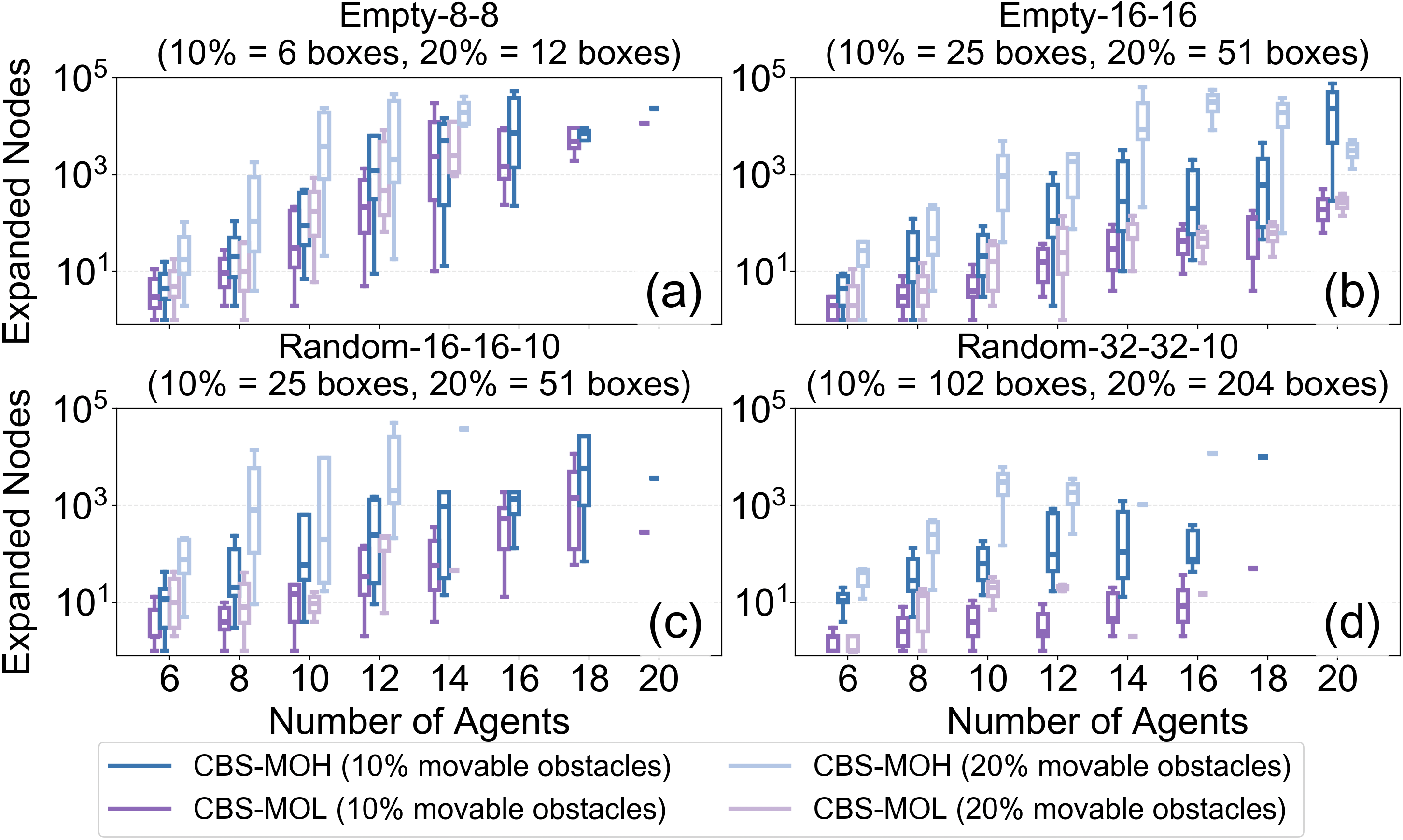}
\caption{Expansions of \abbrOursh and \abbrOursl under a varying number of boxes and a varying number of agents across different grid maps.}
\label{PAMO:fig:expansion}
\end{figure}

Fig.~\ref{PAMO:fig:expansion} reports the number of expanded nodes over instances commonly solved by both \abbrOursh and \abbrOursl, using box plots with medians as central lines. \abbrOursh consistently expands more nodes than \abbrOursl. This is because \abbrOursh employs ST-A* as its low-level planner, which ignores movable obstacles and thus runs faster, allowing more search iterations within the 60-second time limit.
However, by neglecting objects during low-level planning, \abbrOursh often finds more conflicts at the high level, leading to a larger number of node expansions and thus higher search overhead.

To summarize, for a small number of boxes, \abbrOursh is recommended due to its fast running speed per iteration.
For large number of boxes, \abbrOursl may be more advantageous as it considers boxes at low-level.
\FloatBarrier

\FloatBarrier
\subsection{Real-Robot Validation}

\begin{figure}[!t]
\centering
\includegraphics[width=1\linewidth]{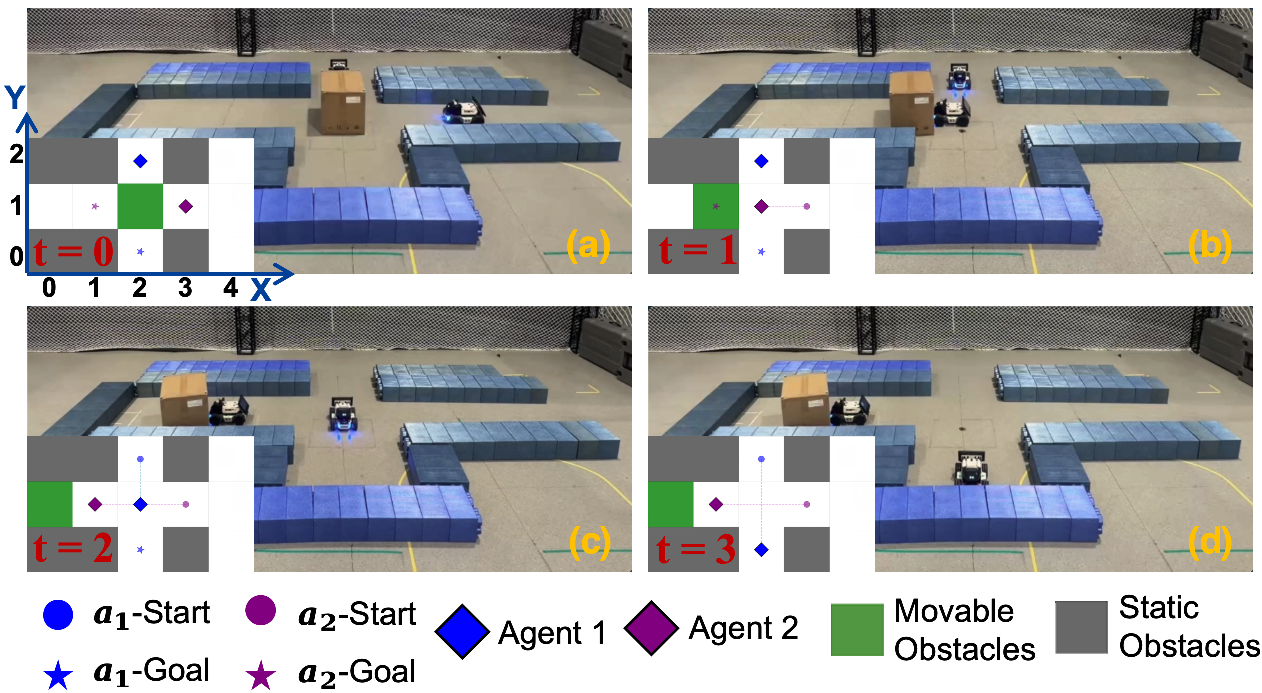}
\caption{An M-PAMO scenario involving two agents and a single movable
obstacle (a box) for path finding. Agent $a_1$ starts at $(2,2)$ with goal $(2,0)$, while agent $a_2$ begins at $(3,1)$ and aims to reach $(1,1)$. The box initially occupies cell $(2,1)$, obstructing the direct path between the agents’ start and goal locations. Subsequently, agent $a_2$ pushes the box away from its original position, thereby clearing the obstruction for $a_1$. Both agents then successfully reach their respective goal locations.}
\label{PAMO:fig:phy_two}
\vspace{-5mm}
\end{figure}

We deploy our planners on real robots (differential-drive LIMO robots) with a motion capture system.
Fig.~\ref{PAMO:fig:phy_two} shows a case where two agents plan paths with a single box in a compact environment.
The purple robot (agent $a_2$) must push the box at first, since otherwise the problem is unsolvable.
Our method is able to find a feasible solution and paths are executed by the robots for verification.

\section{Conclusion and Future Work}\label{dms:sec:conclude}
This paper formulates a new problem \abbrMAPFMO, proposes two approaches to solve it based on the popular CBS method, and compares their performance in various maps.
The complex interaction between the boxes and the agents couples the agents and makes the problem computationally challenging and burdens the popular CBS approach.

For future work, one can improve CBS-based approaches by passing the information about pushed boxes to the low-level, and combining different constraints by adding strategies (e.g., onto agents, onto boxes) for better performance.
Besides, one can also consider actions with different amount of duration where, for example, push actions take longer duration than regular move actions, and study \abbrMAPFMO with asynchronous actions~\cite{ren21lss,2025_AAAI_LSRP_ShuaiZhou,2026_AAMAS_CBSAA_XuemianWu}.



\bibliographystyle{splncs03_unsrt}
\bibliography{ref}

\end{document}